\documentclass{ecai}
\usepackage{graphicx}
\usepackage{latexsym}
\usepackage{hyperref}
\usepackage{url}
\usepackage{amsmath}
\usepackage{amssymb}
\usepackage{multirow}
\usepackage{xcolor}
\usepackage{amsfonts}
\usepackage{comment}
\usepackage{listings}
\newcommand{\name}{\text{ITHN}}

\begin{document}

\begin{frontmatter}

\title{Automatic Radiology Report Generation by Learning with Increasingly Hard Negatives}

\author[A]{\fnms{Bhanu Prakash}~\snm{Voutharoja}}
\orcid{0000-0002-1352-2858}
\author[A]{\fnms{Lei}~\snm{Wang}\thanks{Corresponding Author. Email: leiw@uow.edu.au.}}
\author[B]{\fnms{Luping}~\snm{Zhou}} 

\address[A]{School of Computing and IT, University of Wollongong, NSW, Australia}
\address[B]{School of Electrical and Information Engineering, University of Sydney, NSW, Australia}

\begin{abstract}
Automatic radiology report generation is challenging as medical images or reports are usually similar to each other due to the common content of anatomy. This makes a model hard to capture the uniqueness of individual images and is prone to producing undesired generic or mismatched reports. This situation calls for learning more discriminative features that could capture even fine-grained mismatches between images and reports. To achieve this, this paper proposes a novel framework to learn discriminative image and report features  by distinguishing them from their closest peers, i.e., hard negatives. Especially, to attain more discriminative features, we gradually raise the difficulty of such a learning task by creating increasingly hard negative reports for each image in the feature space during training, respectively. By treating the increasingly hard negatives as auxiliary variables, we formulate this process as a min-max alternating optimisation problem. At each iteration, conditioned on a given set of hard negative reports, image and report features are learned as usual by minimising the loss functions related to report generation. After that, a new set of harder negative reports will be created by maximising a loss reflecting image-report alignment. By solving this optimisation, we attain a model that can generate more specific and accurate reports. 
It is noteworthy that our framework enhances discriminative feature learning without introducing extra network weights.   Also, in contrast to the existing way of generating hard negatives, our framework extends beyond the granularity of the dataset by generating harder samples out of the training set. Experimental study on benchmark datasets verifies the efficacy of our framework and shows that it can serve as a plug-in to readily improve existing medical report generation models. The code is publicly available at \url{https://github.com/Bhanu068/ITHN}



\end{abstract}

\end{frontmatter}

\section{Introduction}
A radiology report is a multi-sentence paragraph that precisely describes the normal and abnormal regions in a radiology image. Writing such reports requires proper experience and expertise \cite{jing-etal-2018-automatic}. Automating this process can reduce manual workload and speed up clinic procedure. Although radiology report generation is similar to image captioning, the former experiences more subtle correlation between medical images and corresponding reports. This could make image captioning models \cite{sat, saat, bottom-up, sct, adatt} fail when directly applied on medical datasets. 
Many works have been proposed for medical report generation recently \cite{jing-etal-2018-automatic, jing-etal-2019-show, li-neurips, chen-etal-2020-generating, chen-etal-2021-cross-modal, self-boosting}, with the focus on improving the generated reports by using medical tags~\cite{jing-etal-2018-automatic}, large pretrain models~\cite{densenet-huang, resnet, gpt2-radford}, relational memory~\cite{chen-etal-2020-generating, chen-etal-2021-cross-modal}, and so on. 

\begin{figure}
    \begin{center}
    \includegraphics[width=.48\textwidth]{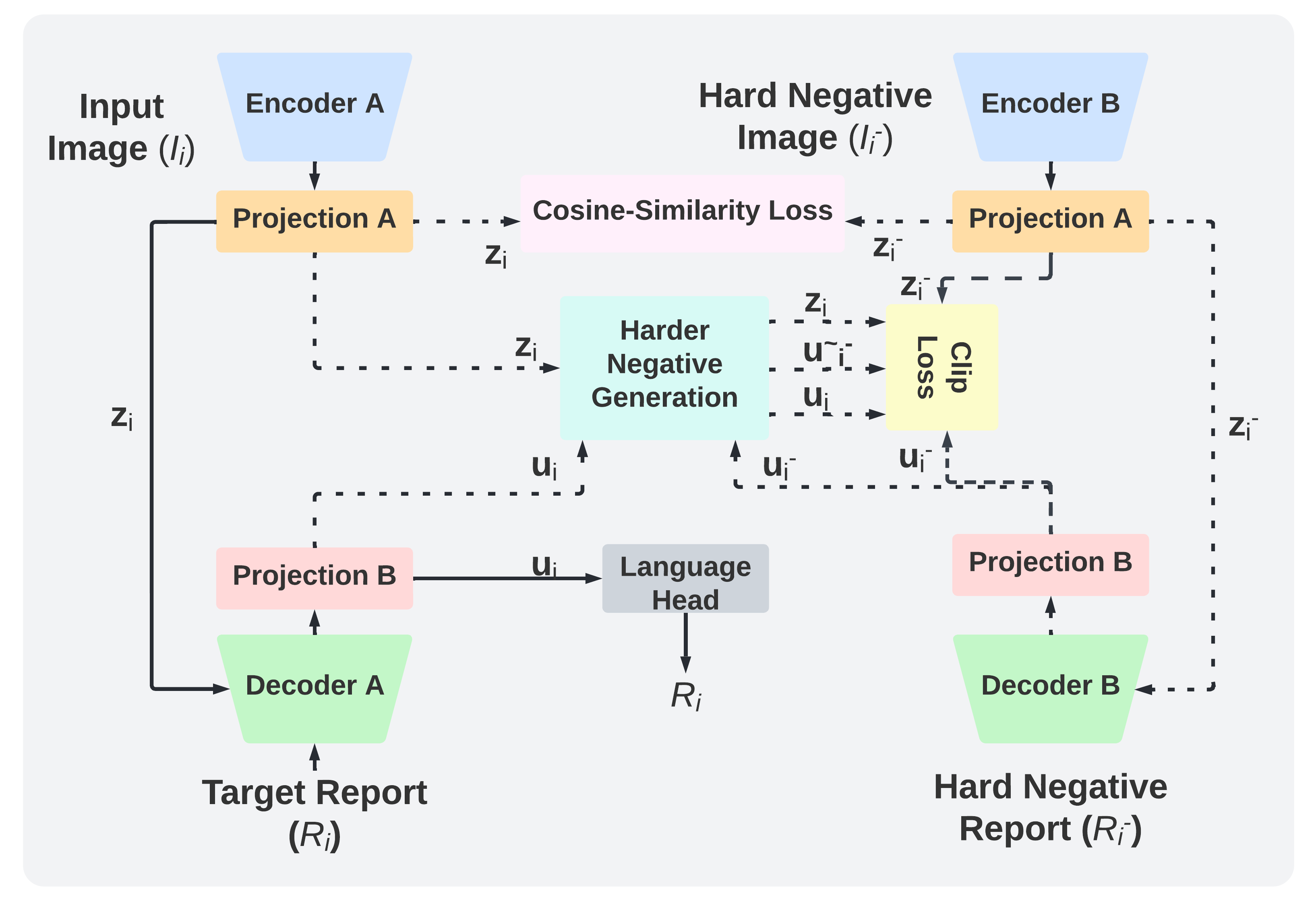}
    \end{center}
    \caption{Based on an encoder-decoder structure, our framework utilises increasingly hard negative reports to enforce the model to learn more distinctive features for both image and report so as to attain accurate and diverse medical reports. At the inference stage, only images are needed as the input for the model to output the generated reports. Detailed explanation is in Sec.~\ref{Sec:proposed}.
    }
    \label{fig:framework}
\end{figure}

Despite of the progress, the generated reports could still be heavily dominated by the terms describing the common contents of medical images \cite{jing-etal-2019-show}, making the characteristics of individual reports submerged. 
Recently, some methods have been proposed to mitigate this issue, for example, by learning sample-specific features for image or report \cite{PPKED} or contrasting the given samples to their closer ones within the training set to guide the report generation \cite{contrastive-attention}.
Nevertheless, they still focus on distinguishing abnormal samples from the common normal ones, which may not be able to sufficiently differentiate the subtle but critical changes across individual reports. Moreover, the two methods solely use either image or text modality to capture abnormality-related features, ignoring the crucial information from the other modality. For example, the former \cite{PPKED} relies on merely text modality, whereas the latter~\cite{contrastive-attention} solely utilizes image modality.

To address this situation, we argue that it is imperative to learn more discriminative features to represent image and report and consider both visual and textual modalities so as to generate reports with sufficient individual characteristics. Following this idea, we propose a method that can better capture the unique features of a given image (or report) by differentiating it from its closest hard negative report (or image) in this paper. Particularly, to adequately capture the fine differences among similar images and reports, we innovatively synthesise \textit{increasingly harder negative reports in the feature space} with the learning process. Specifically, we view the increasingly hard negatives as auxiliary variables and put forward a min-max alternating optimisation to synthesise them during training. At each iteration, by minimisation, we mean to reduce the loss on report generation to update the model weights; By maximisation, we mean to increase the loss on image-report alignment to create a new set of harder negatives to make the "minimisation" in the next iteration more difficult. Through this alternating optimisation, we keep challenging the model to push it to learn more discriminative features such that it can better characterise different medical images and reports and get the true pairs well aligned. After the model is trained, at the inference stage, we only need a medical image as the input to generate a report as usual.

To realise and enrich the above methods, this work also considers the following issues. 
1) When synthesising the increasingly hard negative reports in the learned feature space, we shall maintain their nature, i.e., restricting them to the manifold of ``reports''. We implement this by imposing a simple constraint during the maximisation process. 2) To contrast with our method, we also put forward another way to synthesise increasingly hard negative reports. It is a weighted linear combination of a given report and its initial hard negative in the feature space, which is inspired by a strategy called MoCHi \cite{hard-contrastive}. As will be seen by both theoretical analysis and experimental study, our method shows clear advantage thanks to the optimisation process. 3) We identify the initial set of hard negatives based on the report modality. This is because compared to images, textual reports more directly reflect high-level clinic-related semantics. An initial hard negative can readily be obtained for each report by extracting its features with an off-the-shelf transformer model and identifying its nearest neighbour report in the training set. The anchor report and this identified peer form a "hard negative report pair." 4) We observe that when two reports are nearest neighbours, their ground truth paired images will usually be visually similar. Based on this observation, we regard the two images as nearest neighbours too. In other words, based on the above ``hard negative report pair,'' we form a ``hard negative image pair'' correspondingly. For the two images in a pair, a loss will be  defined in our method to encourage their features to be different from each other. 
5) Lastly, to fully utilise the synthesised harder negatives, we modify the contrastive loss (CLIP) \cite{clip-loss} when using it in our method. A similarity matrix between the features of images and those of the synthesised harder negative reports is additionally computed. It is appended to the original similarity matrix in CLIP loss to form the final matrix to be used.

Our main contributions are summarised as follows.
\begin{itemize}
\item \textit{First}, we propose a simple yet effective method that helps the model to capture discriminative features in both image and text modalities to generate diverse and accurate reports. Our method can readily work with existing encoder-decoder based medical report generation models with a minimum amount of alteration. Moreover, it does not introduce any additional network weights.
\item \textit{Second}, we propose an optimisation-based mechanism to utilise increasingly hard negatives to enhance the learning performance. A principled way is explored to synthesise a set of optimal harder negative reports in the sense of loss function \textit{maximisation}. By training our model with the increasingly harder negatives, the recognition resolution of our model can be increased. Furthermore, unlike the conventional approach of identifying hard negatives from within the data set, our framework extends beyond the granularity of training data by synthesising new, harder negatives.
\item \textit{Third}, we propose an improved variant of the contrastive loss to fully utilise the synthesised hard negative reports rather than merely relying on the negatives randomly sampled into a batch as in the original CLIP loss. 
\end{itemize}
With three different backbones, we validate the proposed framework on two benchmarks IU-XRay~\cite{iu-xray} and MIMIC-CXR~\cite{mimic-cxr} under a variety of assessment criteria. As experimentally demonstrated, it outperforms the existing medical report generation methods and achieves the state-of-the-art performance. 

\section{Related Work}
\label{2}

\noindent \textbf{Medical Report Generation.} Earlier works adopted a traditional encoder-decoder framework for report generation. To generate coherent reports, \cite{jing-etal-2018-automatic} proposed a co-attention mechanism to localize abnormal regions and a hierarchical LSTM decoder to generate the reports. They also adopted disease tag classification task along with report generation, making it a multi-task learning framework. Later, \cite{jing-etal-2019-show} proposed a multi-agent model to alleviate the data bias problem between normal and abnormal regions of the medical images. \cite{chen-etal-2020-generating} incorporated a relational memory module into the vanilla Transformer to store the significant information related to earlier generated reports and utilized this information for effective report generation. Following this work, \cite{chen-etal-2021-cross-modal} used a relational memory matrix for cross-modal alignment of image and text features. After the success of GPT-2 \cite{gpt2-radford} in text generation tasks, \cite{cdgpt2} used GPT-2 for report generation. First, they fine-tuned CheXNet \cite{chexnet} for disease tag prediction, then used the predicted tag's embeddings to calculate weighted semantic features. Finally, they conditioned the GPT-2 model on semantic and visual features to generate a report. \par
To improve the diversity of the generated reports in describing abnormal diseases, \cite{contrastive-attention} proposed contrastive attention mechanism focusing on abnormal regions of the image. It consists of two attentions - aggregate attention to summarise the information from all the reports in the normality pool and differentiate attention to remove common information between normal and abnormal images. Each abnormal image is compared with a set of normal images to get contrastive information. This information is then used for report generation. They try to get the contrastive information from image-level, ignoring the crucial information from text-level.  In other work, PPKED \cite{PPKED} used prior knowledge, posterior knowledge, and multi-domain knowledge distiller to generate the report. This method tries to mimic the behavior of a radiologist, by first predicting the disease tags focusing on the abnormal regions of the image (posterior knowledge), utilising the prior knowledge by retrieving the relevant reports from the corpus, and finally using a distiller module to incorporate the prior and posterior knowledge while doing report generation. Furthermore, \cite{self-boosting} used an additional image-text matching branch to align image and report features in the latent space, and utilised the SBert generated reports during training as the hard negative reports that are close to the ground-truth to supervise branch.\par
The main drawback of these methods is that they try to utilize distinctive information from just one modality. We propose an approach to utilise distinctive features from both modalities and closely align them in the latent space by evolving the hard negatives.

\section{Proposed Method}\label{Sec:proposed}

\subsection{Overview}
\label{3.1}

Given an image set $\mathcal{I} = \{I_{1}, I_{2}, \cdots, I_{n}\}$ and the corresponding ground-truth paired report set $\mathcal{R} = \{R_{1}, R_{2}, \cdots, R_{n}\}$, our framework uses an encoder to extract image representation and passes it to a decoder to generate a report. 

For each report $R_i$, we obtain its nearest neighbour, denoted by $R_i^{-}$, from the training set by using the RadBERT model \cite{radbert} (Sec.~\ref{3.2}). As mentioned in introduction, we also regard their paired images,  denoted by $I_i$ and $I_i^{-}$, as nearest neighbours to each other. 
Images $I_i$ and $I^{-}_i$ are fed into two encoders with shared parameters and reports $R_{i}$ and $R^{-}_{i}$ are fed into two decoders sharing parameters too. The features of the two images extracted by the encoders are denoted as $\mathbf{z}_{i}$ and $\mathbf{z}^{-}_{i}$, respectively. To enforce the encoders to learn features that can differentiate the two neighbouring images, we define a loss to encourage the dissimilarity between $\mathbf{z}_{i}$ and $\mathbf{z}^{-}_{i}$ (Sec.~\ref{3.3}). The two image features are further fed into the decoders to produce the corresponding report features, which are denoted by $\mathbf{u}_{i}$ and $\mathbf{u}^{-}_{i}$, respectively. 
To further enhance the discriminative ability of report features $\mathbf{u}$, we synthesise an increasingly hard negative for each report $\mathbf{u}_{i}$ during the training process, denoted by $\tilde{\mathbf u}_i$, in the feature space (Sec.~\ref{3.4}).
An improved variant of the CLIP loss is then defined to align report feature $\mathbf{u}_{i}$ with the corresponding image feature $\mathbf{z}_{i}$, while pushing the synthesised harder negative features $\tilde{\mathbf u}_i$ away from ${\mathbf z}_i$ (Sec.~\ref{3.5}). Our framework is illustrated in Fig.~\ref{fig:framework}. We elaborate the framework in the following sections. 

\subsection{Identifying Prior Hard Negative Reports}
\label{3.2}

Before training, all reports $\{R_{1}, R_{2}, \cdots, R_{n}\}$ in the training set are fed into RadBERT \cite{radbert}, a pre-trained transformer-based language model tailored to radiology, to extract their report features, denoted by$\{\mathbf{r}_{1}, \mathbf{r}_{2}, \cdots, \mathbf{r}_{n}\}$. For each report ${\mathbf r}_{i}$, its nearest neighbour is identified by a cosine similarity score between ${\mathbf r}_{i}$ and ${\mathbf r}_{j}$, 
where $j = 1,....,n, j \neq i$. By doing so, we create a hard negative report pair $(R_{i}, R^{-}_{i})$ for each report $R_{i}$. Accordingly, we regard their corresponding images as a hard negative image pair $(I_{i}, I^{-}_{i})$ too. 
Note that we create the hard negatives from the perspective of reports as they directly contain high-level clinic-related semantics. The hard negative reports obtained above are used as a prior in our method. Their hardness will be increased gradually during the training, as will be detailed in Sec.~\ref{3.4}.

\subsection{Learning Distinctive Image Features}
\label{3.3}

The pair of images $I_{i}$ and $I^{-}_{i}$ are sent through Encoder $A$ and Encoder $B$ in Fig.~\ref{fig:framework}, respectively, to obtain image features $\mathbf{z}_{i}$ and $\mathbf{z}^{-}_{i}$. 
We then define a loss $\mathcal{L}_{CS}$ to be minimized to make each image $I_{i}$ and its hard negative $I_{i}^{-}$ dissimilar from each other. This aims to learn powerful encoders capable of differentiating the two closely similar images. A common cosine similarity score is used. 
\begin{equation}\label{eqn:L_CS}
\mathcal{L}_{CS} = \frac{\mathbf{z}_{i}\cdot{\mathbf{z}^{-}_{i}}}{\|{\mathbf{z}_{i}}\|\cdot\|{\mathbf{z}^{-}_{i}}\|}.
\end{equation}
In addition, a cross-entropy loss $\mathcal{L}_{CE}$ used to generate reports adds another level of supervision on the encoder to learn image representations useful for report generation. Together, the losses $\mathcal{L}_{CS}$ and $\mathcal{L}_{CE}$ make the image representations from the encoder not only discriminative but also significant for report generation. 

\subsection{Synthesising Increasingly Hard Negative Reports}
\label{3.4}
The image features $\mathbf{z}_{i}$ and $\mathbf{z}^{-}_{i}$ along with the corresponding reports $R_{i}$ and $R^{-}_{i}$ are passed to Decoders $A$ and $B$. The report features produced at the last hidden layer of the two decoders are denoted as $\mathbf{u}_{i}$ and $\mathbf{u}^{-}_{i}$, respectively, where $\mathbf{u}^{-}_{i}$ is the hard negative of $\mathbf{u}_{i}$ as previously defined. The decoders should well differentiate $\mathbf{u}_{i}$ from $\mathbf{u}^{-}_{i}$ so as to achieve discriminative report features. To facilitate this, we argue that the efficacy of decoders could be enhanced by challenging it with increasingly hard $\mathbf{u}^{-}_{i}$, i.e., higher difficulty to differentiate $\mathbf{u}^{-}_{i}$ from $\mathbf{u}_{i}$. To achieve this, two strategies to synthesise harder negative reports are put forward as follows. 

The first strategy is inspired by a hard-negative mixing method MoCHi~\cite{hard-contrastive}. MoCHi designs harder negatives to facilitate better and faster contrastive learning in computer vision tasks such as image classification, object detection, and image segmentation. Following the spirit, we utilise it in our framework for medical report generation. In the feature space, given a report $\mathbf{u}_{i}$ and its $k$ nearest neighbours $\{\mathbf{u}^{-}_{r_1},..., \mathbf{u}^{-}_{r_k}$\} identified via RadBERT before training starts, we synthesise the harder negative, denoted by $\tilde{\mathbf u}^{-}_{i}$, by linearly combining $\mathbf{u}_{i}$ and a randomly sampled $\mathbf{u}^{-}_{r_i}$, where $1\leq{i}\leq{k}$.
\begin{equation}\label{eqn:MoCHi}
    \tilde{\mathbf u}^{-}_{i} = (1 -\lambda)\mathbf{u}^{-}_{r_i} + \lambda\mathbf{u}_{i} 
\end{equation}where $\lambda$ is a parameter controlling the contribution of report $\mathbf{u}_{i}$ to the synthesised harder negative. 

As will be experimentally demonstrated, the harder negatives synthesised by our MoCHi-inspired strategy can indeed improve the quality of the generated reports. However, note that MoCHi~\cite{hard-contrastive} only considers the tasks of single modality (i.e., image-based classification, detection, and segmentation). This nature is also reflected in our MoCHi-inspired strategy in which ${\mathbf u}^{-}_{r_i}$ and $\mathbf{u}^{-}_{i}$ are features from the same modality (i.e., report). Differently, medical report generation involves two modalities, i.e., image and text. Both of them shall be taken into account when synthesising the harder negatives. More importantly, the strategy in Eq.(\ref{eqn:MoCHi}) is disconnected from the training process of report generation and its fitness to our task is not justified. 

The second strategy, which is developed by this work, addresses the above issues by proposing an optimisation-based approach. 
Let ${\mathcal L}_F$ denote our final loss function used to train the model for report generation. We use ${\boldsymbol\theta}$ to denote the set of network weights to be learned and recall that $\tilde{\mathbf u}^{-}_{i}$ stands for the harder negative to be synthesised for a report $\mathbf{u}_{i}$, where $i=1,\cdots,n$.  By treating the harder negatives as auxiliary variables, we can explicitly express ${\mathcal L}_F$ as ${\mathcal L}_F({\boldsymbol\theta};\tilde{\mathbf u}^{-}_{1},\cdots,\tilde{\mathbf u}^{-}_{n})$. To synthesise the harder negatives with the training process, we propose a min-max optimisation problem
\begin{equation}
    \underset{{\boldsymbol\theta}}{\min}\underset{\{\tilde{\mathbf u}^{-}_{1},\cdots,\tilde{\mathbf u}^{-}_{n}\}}{\max}{\mathcal L}_F({\boldsymbol\theta};\tilde{\mathbf u}^{-}_{1},\cdots,\tilde{\mathbf u}^{-}_{n}).
\end{equation}
This optimisation is solved in an alternating manner. At each training iteration, given the current set of $\{\tilde{\mathbf u}^{-}_{1},\cdots,\tilde{\mathbf u}^{-}_{n}\}$, the above problem reduces to $\underset{{\boldsymbol\theta}}{\min}{\mathcal L}_F({\boldsymbol\theta})$, which corresponds to the common case of network weights updating, say, via the back-propagation process. After that, given the updated ${\boldsymbol\theta}$, the above problem reduces to ${\max}_{\{\tilde{\mathbf u}^{-}_{1},\cdots,\tilde{\mathbf u}^{-}_{n}\}}{\mathcal L}_F(\tilde{\mathbf u}^{-}_{1},\cdots,\tilde{\mathbf u}^{-}_{n})$. It means to find a new set of harder negatives that can maximally increase the current loss to make the subsequent learning more challenging.  In report generation tasks, the final loss ${\mathcal L}_F$ is usually a linear combination of multiple sub-loss functions, in which a triplet loss or a contrastive loss depends on the hard negatives while the remaining losses are not.      

In the following, we demonstrate this maximisation problem via a triplet loss. We use this loss by considering that 1) triplet loss will lead to a simpler optimisation for each $\mathbf{u}^{-}_i$; 2) triplet loss makes the comparison of this strategy with the above MoCHi-inspired strategy more clear; and 3) triplet loss shares the similar spirit with the contrastive loss. Let $({\mathbf z}_i,\mathbf{u}_i,\tilde{\mathbf u}^{-}_i)$ denote a triplet, where we recall that ${\mathbf z}_i$ is image feature, $\mathbf{u}_i$ the feature of paired report, and $\tilde{\mathbf u}^{-}_i$ the hard negative to be synthesised. A common triplet loss is expressed as
\begin{equation}
\label{eq:triplet_loss}
    {\mathcal L}(\tilde{\mathbf u}^{-}_i) = \max(\|{\mathbf z}_i-\mathbf{u}_i\|^2-\|{\mathbf z}_i-\tilde{\mathbf u}^{-}_i\|^2+m_0,0),
\end{equation}where $m_0$ is a preset constant of margin. When the first term in the bracket is greater than zero, a loss will be incurred. In this case, the gradient of the loss with respect to $\tilde{\mathbf u}^{-}_i$ can be easily shown as
$\frac{\partial{\mathcal L}(\tilde{\mathbf u}^{-}_i)}{\partial\tilde{\mathbf u}^{-}_i} = 2({\mathbf z}_i-\tilde{\mathbf u}^{-}_i)$.
This indicates that to \textit{increase} the loss, the harder negative shall be synthesised by moving the current $\tilde{\mathbf u}^{-}_i$ towards the image feature ${\mathbf z}_i$.     

\begin{figure}
    \begin{center}
    \includegraphics[width=1.1\linewidth]{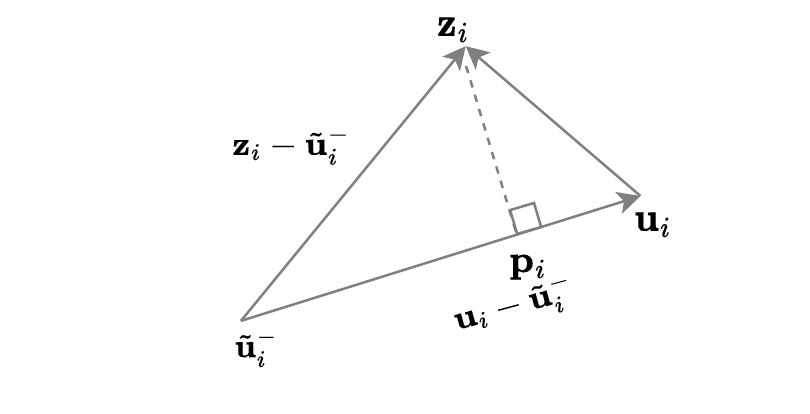}
    \end{center}
    \caption{Illustration of our optimisation-based strategy to synthesize a harder negative report. $\mathbf{z}_{i}$ denotes the feature of image $I_i$, $\mathbf{u}_{i}$ denotes the feature of the paired report $R_i$, $\tilde{\mathbf u}^{-}_{i}$ is the synthesised harder negative in the last iteration. $\mathbf{p}_i$ is the projection of $\mathbf{z}_{i}$ onto the line $\overline{\tilde{\mathbf u}^{-}_{i}\mathbf{u}_{i}}$. The synthesised harder negative is a linear combination of $\tilde{\mathbf u}^{-}_{i}$ and ${\mathbf p}_i$, as shown in Eq.(\ref{eq:eq4}).}
    \label{fig:image_projection}
\end{figure}

Nevertheless, we observe that the harder negatives synthesised in this way degrade the quality of report generation. Our explanation is that this way does not maintain the nature of the synthesised harder negative, i.e., "being a report." To handle this issue, we constrain the synthesised hard negative on the manifold of ``report'' in the feature space. Since the current $\tilde{\mathbf u}^{-}_i$ is the hard negative of $\mathbf{u}_i$, they are close in the feature space.  Therefore, the line segment between the current $\tilde{\mathbf u}^{-}_i$ and $\mathbf{u}_i$, denoted by $\overline{\tilde{\mathbf u}^{-}_{i}\mathbf{u}_{i}}$, will be a good local approximation to the ``report'' manifold near $\tilde{\mathbf u}^{-}_i$. Therefore, we impose the constrain that the synthesised harder negative must reside on the line $\overline{\tilde{\mathbf u}^{-}_{i}\mathbf{u}_{i}}$.  In the presence of such a constraint, we can utilise projected gradient descent to obtain the synthesised harder negative as
\begin{eqnarray} 
    \tilde{\mathbf u}^{-}_i &\triangleq& {\mathcal P}(\tilde{\mathbf u}^{-}_i + \frac{1}{2}\lambda\frac{\partial{\mathcal L}(\tilde{\mathbf u}^{-}_i)}{\partial\tilde{\mathbf u}^{-}_i}) \\~\nonumber
    &=& {\mathcal P}(\tilde{\mathbf u}^{-}_i + \lambda({\mathbf z}_i-\tilde{\mathbf u}^{-}_i)) = {\mathcal P}((1-\lambda)\tilde{\mathbf u}^{-}_i + \lambda{\mathbf z}_i),  
\end{eqnarray}
where ${\mathcal P}({\mathbf x})$ denotes the operator that projects a point ${\mathbf x}$ in the feature space onto the line $\overline{\tilde{\mathbf u}^{-}_{i}\mathbf{u}_{i}}$.  It is not difficult to derive the projection of ${\mathbf z}_i$ onto this line as  
\begin{equation}
    \mathbf{p}_i = \tilde{\mathbf u}^{-}_{i} + \frac{(\mathbf{z}_{i} - \tilde{\mathbf u}^{-}_{i})\cdot (\mathbf{u}_{i} - \tilde{\mathbf u}^{-}_{i})}{||\mathbf{u}_{i} - \tilde{\mathbf u}^{-}_{i}||}  \cdot \frac{\mathbf{u}_{i} - \tilde{\mathbf u}^{-}_{i}}{||\mathbf{u}_{i} - \tilde{\mathbf u}^{-}_{i}||}.
\end{equation}
Formally,  the synthesised harder negative can be expressed as
\begin{equation}\label{eq:eq4}
\tilde{\mathbf u}^{-}_{i} \triangleq (1-\lambda)\tilde{\mathbf u}^{-}_{i} + \lambda\mathbf{p}_i.
\end{equation}
The above process is illustrated in Fig. \ref{fig:image_projection}.

Now we can contrast Eq.(\ref{eq:eq4}) with the MoCHi-inspired strategy in Eq.(\ref{eqn:MoCHi}). It can be seen that both strategies are based on linear combination. Their first terms in the right hand side share the similar nature in that either $\tilde{\mathbf u}^{-}_{i}$ or $\mathbf{u}^{-}_{r_i}$  is a sample close (as hard negative or nearest neighbour) to $\mathbf{u}_{i}$. Meanwhile, through the projection ${\mathbf p}_i$, Eq.(\ref{eq:eq4}) additionally considers the image modality $\mathbf{z}_{i}$, which is missing in the MoChi-inspired one. Finally, it is worth noting that in real implementation, we empirically set $\tilde{\mathbf u}^{-}_i$ in the right hand side of Eq.(\ref{eq:eq4}) as the prior hard negative ${\mathbf u}^{-}_i$ instead of the $\tilde{\mathbf u}^{-}_i$ synthesised at the last iteration to avoid being too aggressive. In this setting, the improvement of our strategy to the MoChi-inspired one will more directly reflect the benefit of considering both image and text modalities when synthesising harder negatives.  


\subsection{Training Process}\label{3.5}
The above process forms an updated triplet $(\mathbf{z}_{i}, \mathbf{u}_{i}, \tilde{\mathbf u}^{-}_{i})$. To capture finer visual findings, highly correlated image and report features are desired. We explicitly align image ($\mathbf{z}_{i}$) and the truly matched report ($\mathbf{u}_{i}$) to be close and push the synthesised harder negative report ($\tilde{\mathbf u}^{-}_{i}$) away from the image. 
To achieve stronger alignment, we use multiple negatives and therefore the contrastive loss, CLIP \cite{clip-loss}, which has a similar spirit to Eq.~(\ref{eq:triplet_loss}) in terms of aligning matched pairs together and pushing unmatched ones away from each other. The CLIP loss for a batch $B$ containing paired sample features $\mathbf{a}$ and $\mathbf{b}$ is   
\begin{equation}
\label{eq:con_loss}
    L_{con} = -\log{\sum_{i=1}^{B}\dfrac{\exp\{sim(\mathbf{a}_{i}, \mathbf{b}_{i}) / \tau\}}{\sum_{k=1; [k \neq i]}^{B} \exp\{sim(\mathbf{a}_{i}, \mathbf{b}_{k}) / \tau\}}}
\end{equation}The loss depends on the similarity matrix of sample features in a batch. The diagonal elements of the matrix are considered to be paired and pulled close to each other, while the off-diagonal elements are unpaired samples and pushed away. 

However, in the CLIP loss, negative samples are usually randomly sampled from a batch of paired samples during training. Due to this, all the negatives within the batch might not really be hard negatives for each positive sample. To address this situation, we propose to compute an additional similarity matrix between image features and the synthesised harder negative report features and then average the two similarity matrices to obtain a final similarity matrix to be used in our CLIP loss.

Specifically, we first calculate a cosine similarity matrix for samples ($\mathbf{z}_{i}$, $\mathbf{u}_{j}$), where $i,j = 1,\cdots,B$ as
\begin{equation}
    {\mathbf S}_1(\mathbf{z}_{i}, \mathbf{u}_{j}) = \frac{{\mathbf{z}_{i}}\cdot{\mathbf{u}_{j}}}{\|{\mathbf{z}_{i}}\|\cdot\|{\mathbf{u}_{j}}\|}. 
\end{equation} A similar matrix ${\mathbf S}_2$ can be obtained in the same way for samples  ($\mathbf{z}^{-}_{i}$, $\mathbf{u}^{-}_{j}$). 
After this, we calculate one more similarity matrix with synthesised harder negative ($\mathbf{z}_{i}$, $\tilde{\mathbf u}^{-}_{i}$), where $i,j = 1,\cdots,B$ as 
\begin{equation}
    {\mathbf S}_3(\mathbf{z}_{i}, \tilde{\mathbf u}^{-}_{j}) =    \frac{{\mathbf{z}_{i}}\cdot{\tilde{\mathbf u}^{-}_{j}}} {\|{\mathbf{z}_{i}}\|\cdot\|{\tilde{\mathbf u}^{-}_{j}}\|}
\end{equation}Meanwhile, since $\tilde{\mathbf u}^{-}_{i}$ is a harder negative to $\mathbf{z}_{i}$, we need to push the diagonal elements of this matrix to be small as the off-diagonal ones do. To implement this, we multiply each of the diagonal elements of this matrix with $-1$ and denote the result matrix as ${\mathbf S}^{'}_{3}$. 

Our similarity matrix is therefore 
\begin{equation}
    {\mathbf S} = \frac{1}{3}({\mathbf S}_1+{\mathbf S}_2+{\mathbf S}^{'}_3).
\end{equation} and it defines the $sim(\cdot,\cdot)$ in Eq.~(\ref{eq:con_loss}).
The CLIP loss in our case is 
\begin{equation}
    \mathcal{L}_{CP} = \frac{1}{2}\left(L_{con}(sim(\cdot,\cdot)) + L_{con}(sim^{\top}(\cdot,\cdot))\right).
\end{equation}
For report generation, we use the standard cross-entropy loss $\mathcal{L}_{CE}$. By also considering the loss $\mathcal{L}_{CS}$ in Eq.~(\ref{eqn:L_CS}), our final learning objective is defined as
\begin{equation}
\label{eq:eq8}
\mathcal{L}_{F} = \mathcal{L}_{CE} + \beta \mathcal{L}_{CP} + \gamma \mathcal{L}_{CS}.
\end{equation}The hyper-parameters $\beta$ and $\gamma$ balance different objective terms.

\begin{table*}[h]
\begin{center}
\begin{tabular}{lllllllll}
\hline\hline
Dataset & Methods & B-1 & B-2 & B-3 & B-4 & R-L & M & C  \\
\hline\hline
\multirow{14}{*}{IU-XRay\cite{iu-xray}} & HRGR$^{*}$\cite{li-neurips} & 0.438 & 0.298 & 0.208 & 0.151 & 0.322 & - & -\\
& CO-ATT$^{*}$\cite{jing-etal-2018-automatic} & 0.455 & 0.288 & 0.205 & 0.154 & 0.369 & - & -\\
& CMAS-RL$^{*}$\cite{jing-etal-2019-show} & 0.464 & 0.301 & 0.210 & 0.154 & 0.362 & - & -\\
& R2Gen$^{***}$\cite{chen-etal-2020-generating} & 0.458 & 0.295 & 0.210 & 0.159 & 0.375 & 0.176 & 0.408\\
& MedSkip$^{**}$\cite{medskip} & 0.467 & 0.297 & 0.214 & 0.162 & 0.355 & 0.187 & -\\
& KERP$^{**}$\cite{kerp} & 0.470 & 0.304 & 0.219 & 0.165 & 0.371 & 0.187 & 0.280\\
& PPKED$^{**}$\cite{PPKED} & 0.483 & 0.315 & 0.224 & 0.168 & 0.376 & 0.190 & 0.351\\
& CA$^{**}$\cite{contrastive-attention} & 0.492 & 0.314 & 0.222 & 0.169 & 0.381 & 0.193 & -\\
\cline{2-9}
& R2GenCMN$^{***}$\cite{chen-etal-2021-cross-modal} & 0.467 & 0.300 & 0.210 & 0.156 & 0.369 & 0.185 & 0.401\\
& R2GenCMN + \textbf{MoCHi} & 0.489 & 0.310 & 0.211 & 0.151 & 0.374 & 0.190 & 0.440\\
& R2GenCMN + \textbf{\name} \textbf{(ours)} & \bf{0.496} & 0.319 & 0.220 & 0.160 & 0.378 & 0.194 & 0.461\\\cline{2-9}
& XproNet$^{***}$\cite{xpronet} & 0.468 & 0.299 & 0.205 & 0.158 & 0.367 & 0.181 & 0.375\\
& XproNet + \textbf{MoCHi} & 0.455 & 0.306 & 0.224 & 0.169 & 0.382 & 0.196 & 0.410\\
& XproNet + \textbf{\name} \textbf{(ours)} & 0.491 & \bf{0.328} & \bf{0.231} & \bf{0.183} & \bf{0.387} & \bf{0.210} & \bf{0.495}\\
\hline\hline
\multirow{9}{*}{MIMIC-CXR\cite{mimic-cxr}} & CA$^{**}$\cite{contrastive-attention} & 0.350 & 0.219 & 0.148 & 0.106 & 0.278 & 0.142 & - \\
& R2Gen$^{***}$\cite{chen-etal-2020-generating} & 0.363 & 0.216 & 0.143 & 0.101 & 0.269 & 0.135 & 0.141\\
& PPKED$^{**}$\cite{PPKED} & 0.360 & 0.224 & 0.149 & 0.106 & 0.284 & 0.149 & 0.237 \\\cline{2-9}
& R2GenCMN$^{***}$\cite{chen-etal-2021-cross-modal} & 0.369 & 0.223 & 0.148 & 0.105 & 0.270 & 0.136 & 0.143\\
& R2GenCMN + \textbf{MoCHi} & 0.364 & 0.225 & 0.152 & 0.109 & 0.276 & 0.139 & 0.237\\
& R2GenCMN + \textbf{\name} \textbf{(ours)} & \bf{0.375} & 0.226 & \bf{0.155} & 0.112 & 0.284 & 0.142 & 0.300\\\cline{2-9}
& XproNet\cite{xpronet} & 0.344 & 0.215 & 0.146 & 0.105 & 0.279 & 0.138 & 0.359\\
& XproNet + \textbf{MoCHi} & 0.354 & 0.220 & 0.150 & 0.112 & 0.283 & 0.138 & 0.371\\
& XproNet + \textbf{\name} \textbf{(ours)} & 0.370 & \bf{0.230} & 0.154 & \bf{0.114} & \bf{0.294} & \bf{0.145} & \bf{0.387}\\\cline{2-9}
\hline\hline
\multirow{3}{*}{COCO\cite{coco}} & VisualGPT$^{***}$\cite{visualgpt} & 0.677 & - & - & 0.236 & 0.486 & 0.221 & 0.768\\
& VisualGPT + \textbf{MoCHi} & 0.694 & - & - & 0.250 & 0.493 & 0.226 & 0.830\\
& VisualGPT + \textbf{\name} \textbf{(ours)} & \bf{0.701} & - & - & \bf{0.259} & \bf{0.516} & \bf{0.235} & \bf{0.852}\\
\hline\hline
\end{tabular}
\caption{Comparison of our approach and SOTA models on IU-XRay, MIMIC-CXR, and COCO datasets. * indicates the results quoted from \cite{chen-etal-2020-generating} for IU X-Ray and MIMIC-CXR datasets. ** indicates the results taken directly from the respective paper. *** indicates the results obtained by running the author-released codes using the same dataset split as our approach. The metrics B-k (k = 1 to 4) is for BLEU, R-L for ROUGE-L, M for METEOR, and C for CIDER.}
\label{tab:table1}
\end{center}
\end{table*}

\section{Experimental Results}
\label{4}
\subsection{Datasets}
\label{4.1}
We evaluate our framework on two radiology benchmarks IU-XRay \cite{iu-xray} and MIMIC-CXR \cite{mimic-cxr} for radiology report generation. In addition, we show the generality of our framework on the image captioning benchmark COCO \cite{coco}. 

\textbf{IU-XRay} \cite{iu-xray} is a classic radiology dataset from Indiana University with 7,470 frontal and/or lateral X-ray images and 3,955 radiology reports. Each report consists of impressions, findings, and indication sections. The findings section contains multi-sentence paragraphs describing the image and is used as the ground-truth, following the previous works \cite{li-neurips, jing-etal-2019-show, chen-etal-2020-generating, chen-etal-2021-cross-modal}.

\textbf{MIMIC-CXR} \cite{mimic-cxr} is the large-scale radiology dataset having 377,110 images with 227,835 reports from 64,588 patients. We use the official data split with 368,960 training samples, 2,991 validation samples, and 5,159 test samples.

\textbf{COCO} \cite{coco} is the most widely used and standard dataset for image captioning. It comprises 120,000 images, each with 5 different captions. We used the split provided by Karapathy~\cite{Karpathy2017DeepVA}, where 5,000 images are used for validation, 5,000 images for testing, and the rest of the images for training. We download the split COCO dataset from the GitHub repository of the meshed-memory-transformer\footnote{https://github.com/aimagelab/meshed-memory-transformer}. 

\indent For IU-XRay, samples without complete findings sections are removed, following~\cite{li-neurips}. The filtered images and reports for IU-XRay and MIMIC-CXR are publicly available in this repository\footnote{https://github.com/cuhksz-nlp/R2Gen} by \cite{chen-etal-2020-generating}, and they are directly downloaded and used in all our experiments.

\begin{table}
\centering
\begin{tabular}{l|c|c|c|c}
Model & P & R & F1 & D\\\hline
R2Gen\cite{chen-etal-2020-generating} & 0.288 & 0.215 & 0.246 & 0.50\\
R2GenCMN\cite{chen-etal-2021-cross-modal} &  0.295 & 0.221 & 0.252 & 0.55\\
XproNet\cite{xpronet} &  0.302 & 0.222 & 0.255 & 0.64\\
R2GenCMN + \textbf{MoCHi} &  0.321 & 0.252 & 0.282 & 0.63\\
R2GenCMN + \textbf{\name} &  0.355 & 0.281 & 0.313 & 0.69\\
XproNet + \textbf{MoCHi} &  0.333 & 0.265 & 0.295 & 0.71\\
XproNet + \textbf{\name} & \textbf{0.362} & \textbf{0.283} & \textbf{0.317} & \textbf{0.82}\\\hline
\end{tabular}
\caption{Comparison of clinical metrics and diversity scores on MIMIC-CXR. P is precision, R is recall, F1 is F1 score, and D is diversity score.}
\label{tab:ce_metrics}
\end{table}

\subsection{Implementation Details}
\label{4.3}

We test our framework with different encoder-decoder backbones, including R2GenCMN~\cite{chen-etal-2021-cross-modal}, which is our baseline model, and XproNet~\cite{xpronet}, a SOTA model for generating medical reports. Additionally, we integrate our framework into VisualGPT~\cite{visualgpt}, a SOTA image captioning model. We use PyTorch to build our framework. Each input image is resized to $3 \times 256 \times 256$ and later randomly cropped to the size of $3 \times 224 \times 224$. We do not combine the frontal and lateral views of the image and rather pass them separately. The baseline model is trained using Adam \cite{adam} optimizer with a weight decay of $5 \times 10 ^ {-5}$ for 50 epochs, and a batch size of 16. The hidden dimensions for the image and report features are 2048 and 512, respectively. To achieve alignment between the image and report features in a shared space, we employ a linear layer with a hidden size of 512, followed by an l2-normalization layer for both the image and report features. This ensures that both sets of features are projected into the common embedding space. The loss weights in Eq.(\ref{eq:eq8}) are set as $0.1$ and $0.2$ for $\beta$ and $\gamma$, respectively, for all datasets. The initial learning rate is set as $1 \times 10^{-4}$ and further reduced by $10$ times if there is no improvement in either BLEU-3 or BLEU-4 score on the validation set. The value of $\lambda$, in Eq. (\ref{eqn:MoCHi}) and Eq. (\ref{eq:eq4}) is equally set as $1 - e^{-\alpha*\#epoch}$, where $\alpha$ is a hyperparameter and \#epoch denotes the current epoch number. As the epoch number increases, the value of $\lambda$ gradually increases from 0. Our experimental evaluations reveal that the value of $\alpha=0.01$ leads to superior performance, as shown in Table \ref{tab:table1}. When evaluating the existing SOTA models with or without adding our proposed framework, we use default hyper-parameters, optimizer, scheduler, and experimental settings of their methods for a fair comparison. We performed hyperparameter analysis only for $\alpha$, $\beta$, and $\gamma$ values. We have chosen the best values based on the higher BLEU-4 score on the validation set. We  use one NVIDIA V100 32GB GPU for all our experiments. The training took nearly 3 hours to complete for IU X-Ray and 3 days for MIMIC-CXR. \\

\noindent \textbf{Evaluation Metrics}: Following the previous works in medical report generation and image captioning, we evaluate our model on widely used Natural Language Generation (NLG) metrics: BLEU \cite{bleu}, CIDER \cite{cider}, METEOR \cite{meteor} and ROUGE-L \cite{rouge}. The evaluation metrics are readily available in pycocotools \footnote{https://github.com/tylin/coco-caption} package and we directly use it for our experiments. In addition, following the previous works \cite{chen-etal-2020-generating,chen-etal-2021-cross-modal,xpronet}, we also evaluate our framework on clinical efficacy (CE) metrics. To calculate CE metrics, the model trained on CheXpert~\cite{chexpert} is used to label the generated reports and compare the findings to ground-truth reports in 14 distinct categories pertaining to thoracic disorders and support equipment. To assess model performance for CE metrics, we employ {\it precision}, {\it recall}, and {\it F1}. Furthermore, to measure the uniqueness of the generated reports, we use a metric called {\it diversity}, proposed by a recent work SimCTG \cite{simctg}

\subsection{Results}
\label{4.4}

\begin{table*}
\begin{center}
\begin{tabular}{llll||lllll}
\hline
Model (tested on IU X-Ray) & B-4 & R-L & C  & Dataset & $\alpha$ & B-4 & R-L & C \\
\hline
XproNet & 0.158 & 0.367 & 0.375 & \multirow{3}{*}{IU-XRay} & 0.1 & 0.182 & 0.384 & 0.488\\
XproNet + $\mathcal{L}_{con}$ & 0.160 & 0.371 & 0.398 & & 0.01 & \textbf{0.183} & \textbf{0.387} & \textbf{0.495}\\
XproNet w/o harder negative synthesis & 0.165 & 0.375 & 0.426 & & 0.001 & 0.177 & 0.379 & 0.482\\\cline{5-9}
XproNet + \textbf{MoCHi} & 0.169 & 0.382 & 0.410 &\multirow{3}{*}{COCO} & 0.1 & \bf{0.259} & \bf{0.516} & \bf{0.852} \\
XproNet + \textbf{\name} & \textbf{0.183} & \textbf{0.387} & \textbf{0.495} && 0.01 & 0.256 & 0.499 & 0.849 \\
XproNet + \textbf{\name} w/o $\mathcal{L}_{CS}$ & 0.178 & 0.384 & 0.477 && 0.001 & 0.252 & 0.492 & 0.838\\\hline
\end{tabular}
\caption{Ablation studies on component contributions (left) and the impact of hyper-parameter $\alpha$. B is for BLEU, R-L for ROUGE-L, and C for CIDER.}
\label{tab:table-all}
\end{center}
\end{table*}

We first compare our model with the SOTA medical report generation models CO-ATT~\cite{jing-etal-2018-automatic}, CMAS-RL~\cite{jing-etal-2019-show}, HRGR~\cite{li-neurips}, R2Gen~\cite{chen-etal-2020-generating}, R2GenCMN~\cite{chen-etal-2021-cross-modal}, PPKED~\cite{PPKED}, KERP~\cite{kerp}, XproNet~\cite{xpronet}, MedSkip~\cite{medskip}, and CA~\cite{contrastive-attention}. Since the models PPKED, KERP, MedSkip, and CA are not open-sourced, we directly quote the results published in their literature. For the open-sourced models R2Gen\cite{chen-etal-2020-generating}, R2GenCMN~\cite{chen-etal-2021-cross-modal}, and XproNet~\cite{xpronet},  we run their codes with their default experimental settings. Moreover, to further verify if our framework also benefits generic image captioning tasks, we also incorporate our framework into the image captioning model VisualGPT~\cite{visualgpt} and test it on the image captioning benchmark COCO.

\subsubsection{Quantitative Analysis}
\label{4.4.1}
Table \ref{tab:table1} demonstrates the effectiveness of our approach compared with the SOTA methods. We call our approach ITHN, short for image-text hard negatives. All the results with \name{} in Table \ref{tab:table1} are averages of three runs with three random seed values. After integrating our framework into the baseline and the SOTA models in radiology report generation and generic image captioning, we consistently observe a significant boost in the scores of evaluation. The best performance is mostly achieved when incorporating our framework into the strong backbones of the SOTA methods, showing that our way of improvement is orthogonal to the current efforts seen in the literature, and could further benefit the latter. On comparing the results of our approach \name{} with that of MoCHi, we could observe an average relative increase of +6.6\% in BLUE, +7.1\% in METEOR, and +20\% in CIDER for the best performing model XproNet on IU-XRay dataset. These scores indicate that utilising both image and text modalities is an effective approach to synthesising harder negatives, as opposed to relying solely on one modality as in the case of MoCHi. Moreover, mining the hard negatives in our way could also enhance the standard image captioning on the COCO dataset. Our approach significantly boosts the performance of the recently proposed VisualGPT across all NLG metrics. It is noteworthy that out of all NLG metrics, our increase in CIDER scores is the most significant across all the models. CIDER in general could reflect the diversity of the generated text as it down-weights the common content across reports. Moreover, we observe consistent improvement over SOTA models even in clinical and diversity metrics shown in Table \ref{tab:ce_metrics}.

\begin{figure*}[h]
    \begin{center}
    \includegraphics[width=1.\textwidth]{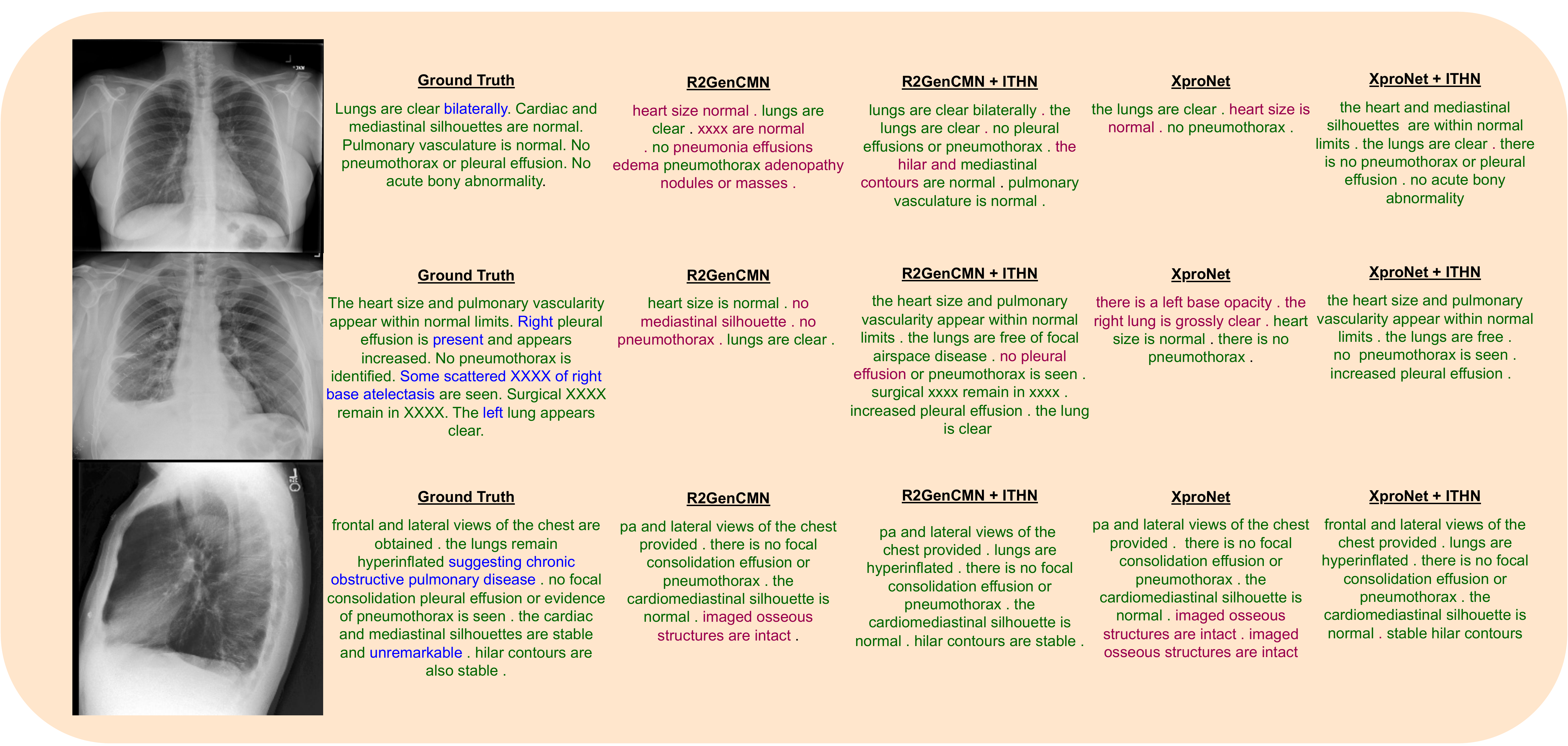}
    \caption{Examples of reports generated by various models. From top to bottom, images are taken from the IU-XRay. The green colour represents the text semantically similar to the ground-truth. The red colour represents the text incorrectly generated. The blue colour in the ground-truth report represents the text not generated by any of the models under comparison.
    }
    \label{fig:comparison}
    \end{center}
\end{figure*}
\subsubsection{Qualitative Analysis}
\label{4.4.2}
In Fig. \ref{fig:comparison}, we compare the reports generated by the SOTA models with and without using our learning framework. The generated sentences are differently colored, with green indicating semantically correct generation, red indicating incorrect generation, and blue indicating the text present in the ground-truth but not generated by any models under comparison. As can be seen, comparing the reports generated by the baseline R2GenCMN and XproNet, incorporating our framework could reduce the incorrect sentences generated in the reports. For instance, in the example from row 3 of Fig. \ref{fig:comparison}, the SOTA models trained with our framework can identify and generate the sentences - "hilar contours are stable" and "lungs are hyperinflated", which are missing in the generated reports without using our framework. This is consistent with the quantitative analysis.

\subsection{Ablation Study} 
\label{4.5}
To understand the contribution of each component of our framework, we conduct an ablation study on three evaluation metrics - BLEU-4, ROUGE-L, and CIDER on IU-XRay.  We use R2GenCMN as our baseline. The results are shown in Table~\ref{tab:table-all} (left). Firstly, to show the impact of contrastive loss on model performance, we add $\mathcal{L}_{con}$ to XproNet. To show the importance of increasing the hardness of reports during training, we compare the results of "XproNet w/o harder negative synthesis" with "XproNet + \textbf{MoCHi}", and "XproNet + \textbf{\name}". We observe that synthesising harder negatives can improve the results by +14.3\% on BLEU, +4.3\% on Rouge-L , and +24.3\% on CIDER. These improvements indicate that the model learns more fine-grained characteristics between the image and report by making the prior hard negative reports much harder during training. Finally, after removing the cosine-similarity loss $\mathcal{L}_{CS}$ in Eq. (\ref{eq:eq8}), which supervises the encoder to capture distinctive image features, we could see a drop in performance. This supports our motivation of learning distinctive features from both modalities to achieve better performance. Moreover, to show the impact of $\alpha$ which controls the rate of hardness increase of the synthesised harder negative features computed using Eq. (\ref{eq:eq4}), we investigate the performance of the model with different values of $\alpha$ as in Table \ref{tab:table-all} (right). The optimal value of $\alpha$ depends on the nature of the dataset. The higher the $\alpha$ value, the faster the rate of increase in the hardness of negative reports will be. For medical datasets such as IU-Xray and MIMIC-CXR, where most medical images have reports of similar nature, having a lower $\alpha$ value gives better performance. On the other hand, for a diverse dataset such as COCO, a slightly higher $\alpha$ value gives better results.

\section{Conclusion}
\label{5}
This paper proposes a framework to improve medical report generation by learning significant and unique features for images and reports. We demonstrate that our framework can readily work with a variety of SOTA medical report generation models, and consistently boost the performance of the latter. We also show that even generic image captioners could benefit from our framework although this is not the focus of this paper. Meanwhile, in terms of potential risk, the medical reports generated by current AI have not achieved human-level performance and therefore shall not be used for practical diagnosis at this stage.



\bibliography{ecai}

\end{document}